\documentclass[runningheads]{llncs}

\usepackage{accv}

\usepackage{accvabbrv}
\usepackage{graphicx}
\usepackage{float}
\usepackage{booktabs}
\usepackage{amsmath,amssymb}
\usepackage{array}
\usepackage{tabularx}
\usepackage{makecell}
\usepackage{adjustbox}
\usepackage{url}
\usepackage[accsupp]{axessibility}

\usepackage[breaklinks,colorlinks,citecolor=accvblue,linkcolor=accvblue,urlcolor=accvblue]{hyperref}

\usepackage{orcidlink}

\begin{document}
\setlength{\emergencystretch}{2em}

\title{KoUniTalk: A Lightweight Articulation-Centered\\
Korean-English 3D Talking Face Benchmark}
\titlerunning{KoUniTalk: A Lightweight Articulation-Centered Benchmark}

\author{Hyunjung Chung \and Unsang Park\thanks{Corresponding author.}}
\authorrunning{H. Chung and U. Park}
\institute{Sogang University, Seoul, Republic of Korea\\
\email{\{phone1263,unsangpark\}@sogang.ac.kr}}

\maketitle

\begin{abstract}
High-quality 3D talking face datasets remain largely English-centric, and Korean
3D facial motion data are difficult to combine with standard English benchmarks
because of differences in mesh topology, spatial scale, coordinate system, and
temporal sampling.
We present \emph{KoUniTalk}, a lightweight articulation-centered Korean-English
3D talking face benchmark that retargets VOCASET and the released Korean
speech-based 3D talking face data to a shared mesh topology using deformation
transfer.
Rather than proposing a new deformation-transfer algorithm or a full-head
identity-preserving avatar dataset, KoUniTalk provides an identity-neutral canonical
output space for controlled speech-driven facial articulation training and evaluation
across English and Korean.
The unified template contains 1,176 vertices and focuses on the mouth and adjacent
lower- and mid-face regions, reducing the output dimensionality from 15,069 and
72,147 dimensions to 3,528 dimensions, corresponding to 4.27-fold and 20.45-fold
reductions compared with VOCASET/FLAME and the original Korean mesh,
respectively.
To examine whether retargeting preserves speech-relevant motion, we evaluate semantic
mouth-landmark trajectories, including mouth opening, mouth width, aperture ratio,
and mouth-opening dynamics.
Since the official test set of the Korean dataset is not publicly released, we
additionally define a subject-disjoint Korean benchmark split.
The processed matched benchmark contains 22 speakers, 4,978 sequences, and 642,781
frames, enabling Korean-English cross-dataset evaluation of speech-driven 3D facial
animation models in a single compact articulation-template space. Source-reported
inventory counts are listed separately from these processed counts.
\end{abstract}

\section{Introduction}
\label{sec:intro}

3D talking face animation has attracted considerable attention as a key technology in
diverse applications such as virtual avatars, AI news anchors, and communication
assistive systems.
Deep learning-based methods, including MeshTalk~\cite{richard2021meshtalk},
FaceFormer~\cite{fan2022faceformer}, and Karras et al.~\cite{karras2017audio}, have
demonstrated promising results in generating 3D talking meshes directly from speech,
yet they require large-scale, high-quality 3D datasets.

Most publicly available datasets are centered on English speakers, resulting in a
significant lack of linguistic diversity.
For languages such as Korean, differences in phoneme--viseme distributions and
articulation patterns can introduce domain gaps that are not well covered by
English-centric benchmarks.
Accordingly, target-language 3D speech data and compatible evaluation protocols are
important for studying multilingual talking face models.

The primary barrier to integrating heterogeneous datasets is mesh topology mismatch.
Datasets with differing vertex counts and face connectivity structures cannot be
directly merged; VOCASET employs triangular meshes while the released Korean speech-based data uses quadrilateral meshes,
making the mesh structures fundamentally incompatible.
Resolving this requires a mesh alignment pipeline to a common template.
Furthermore, the approximately 20-fold vertex density gap between the released Korean speech-based data (24,049 vertices)
and our unified template (1,176 vertices) demands careful parameter design.

\begin{figure}[t]
  \centering
  \includegraphics[width=0.85\linewidth]{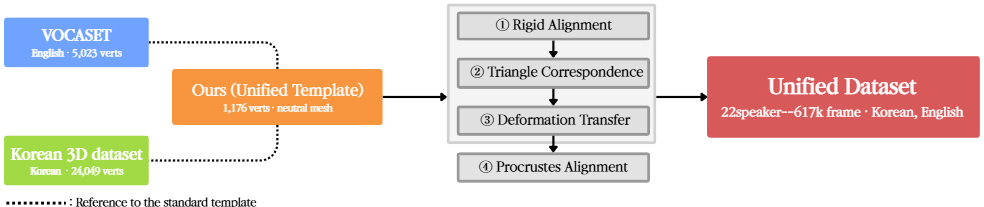}
  \caption{Dataset integration pipeline. Starting from two heterogeneous source
           datasets (VOCASET and the released Korean speech-based data), we apply Deformation Transfer to a
           common unified template and then perform Procrustes Alignment to produce
           a coordinate-consistent multilingual dataset.}
  \label{fig:pipeline}
\end{figure}

In this paper, we integrate VOCASET~\cite{cudeiro2019vocaset} and the released Korean speech-based 3D talking face data~\cite{aihub2023} into a single output space using a custom lightweight unified
template.
Since the two source datasets differ substantially in scale and orientation, we first
coarsely align the unified template to each source dataset manually, then apply
Deformation Transfer~\cite{sumner2004deformation} to propagate per-frame facial
motions onto the unified template.
Finally, because the two resulting mesh sets still exhibit residual scale and
orientation discrepancies, we apply Procrustes Alignment to bring them into a common
coordinate frame.
The overall workflow is illustrated in Fig.~\ref{fig:pipeline}.
The resulting dataset is aligned to a single topology and can be used to configure
Korean-English training and evaluation environments for existing speech-driven 3D
facial animation models. KoUniTalk should be interpreted as an articulation-centered
benchmark rather than a full-head identity-preserving avatar benchmark: it uses an
identity-neutral template to study canonical speech-related mouth and lower/mid-face
motion across languages.

Our contributions are summarized as follows:
\begin{itemize}
  \item We construct a unified-topology Korean-English 3D talking face benchmark by
        retargeting VOCASET and the released Korean speech-based 3D talking face data to a common
        lightweight articulation template.
  \item We introduce an articulation-centered lightweight representation that reduces
        output dimensionality while retaining the mouth and adjacent lower/mid-face
        geometry required for speech-driven articulation analysis.
  \item We define a subject-disjoint Korean evaluation split for the released Korean
        subset and introduce semantic mouth-landmark metrics to evaluate how well the
        retargeting process preserves speech-related mouth articulation.
\end{itemize}

\section{Related Work}
\label{sec:related}

\subsection{Speech-Driven 3D Facial Animation}

Speech-driven 3D facial animation has progressed from early end-to-end audio-driven
animation models~\cite{karras2017audio,taylor2017speech} to neural mesh generation
methods that learn speech-to-vertex mappings from 4D face data.
VOCA~\cite{cudeiro2019vocaset} introduced a speaker-conditioned model for learning and
synthesizing 3D speaking styles. MeshTalk~\cite{richard2021meshtalk} addressed
cross-modality disentanglement between audio and facial motion.
Recent architectures further improve temporal modeling and lip accuracy using
transformers~\cite{fan2022faceformer}, discrete motion priors~\cite{xing2023codetalker},
self-supervised cross-modal training~\cite{peng2023selftalk}, key-motion
embeddings~\cite{xu2024kmtalk}, 2D-to-3D supervision~\cite{zhuang2024learn2talk},
probabilistic motion modeling~\cite{yang2024probabilistic}, and diffusion-based
generation~\cite{stan2023facediffuser,sun2024diffposetalk,lin2025glditalker}.
In parallel, recent works explicitly target lip readability and speech articulation
through audio-visual guidance or phonetic/viseme-aware objectives~\cite{han2024avguide,kim2025viseme}.
Other works emphasize emotional expression, personalization, and speaker
style~\cite{peng2023emotalk,thambiraja2023imitator,kim2025memorytalker}.
Related methods address regional facial motion control and holistic
expression--gesture generation~\cite{wu2023composite,chen2024diffsheg}.
Recent scaling efforts further highlight the need for multilingual data and unified
training across heterogeneous annotations~\cite{sungbin2024multitalk,fan2024unitalker}.
These methods demonstrate substantial progress in speech-driven facial animation, but
most are trained and evaluated within a fixed output topology or require model-side
multi-head handling for different annotation spaces. KoUniTalk addresses a
complementary problem: constructing a common articulation-centered output space that
enables controlled Korean-English evaluation across originally incompatible mesh
datasets.

\subsection{3D Talking Face Datasets and Language Coverage}

VOCASET~\cite{cudeiro2019vocaset} consists of approximately 104,400 frames recorded
from 12 English speakers, providing 5,023-vertex sequences based on the FLAME
template at 60\,fps, and has been widely adopted as a benchmark in works such as
FaceFormer~\cite{fan2022faceformer} and CodeTalker~\cite{xing2023codetalker}.
BIWI~\cite{fanelli2010biwi} and MultiFace~\cite{wuu2022multiface} provide additional
3D facial motion resources, but public speech-driven 3D benchmarks remain heavily
English-centric and are usually tied to their own mesh representations.
Broader audio-visual talking-face resources, including AVSpeech, MEAD, and HDTF,
support large-scale video-based speech or talking-face research~\cite{ephrat2018looking,wang2020mead,zhang2021hdtf},
and OLKAVS provides a large-scale Korean audio-visual speech resource~\cite{park2024olkavs}.
However, these resources do not directly provide Korean-English 3D mesh sequences in
a shared audio-synchronized topology.
Large-scale and dynamic 3D face resources such as FaceWarehouse, FaceScape, 4DFAB,
and D3DFACS provide valuable identity, expression, and motion geometry~\cite{cao2014facewarehouse,yang2020facescape,cheng20184dfab,cosker2011d3dfacs},
but they are not designed as Korean-English speech-driven 3D talking face benchmarks
with a shared output topology.
The released Korean speech-based 3D talking face data~\cite{aihub2023} is a large-scale Korean-specific
dataset comprising 512,780 frames from 10 Korean speakers, providing 24,049-vertex
sequences at 30\,fps based on its own template.
However, because its topology, vertex density, and coordinate conventions differ from
VOCASET/FLAME, it cannot be directly concatenated with existing English 3D talking
face benchmarks. Our work therefore targets the integration of VOCASET's English
sequences and the Korean sequences into a shared articulation-template space.

\subsection{Mesh Retargeting and Canonical Face Spaces}

Deformation Transfer proposed by Sumner and Popović~\cite{sumner2004deformation} is a
deformation propagation technique applicable even when vertex counts, face counts, and
connectivity differ between meshes.
It consists of correspondence establishment followed by per-frame deformation transfer:
the correspondences are computed once per source--target pair and then reused across
all frames of the corresponding sequence.
Classical registration methods, including rigid ICP~\cite{besl1992method},
Coherent Point Drift~\cite{myronenko2010point}, and non-rigid ICP~\cite{amberg2007nonrigid},
aim to align heterogeneous 3D surfaces to a common template, while deformation
regularizers such as as-rigid-as-possible modeling~\cite{sorkine2007arap} provide
useful priors for surface deformation.
Classical morphable face models~\cite{blanz1999morphable}, FLAME~\cite{li2017flame},
and mesh autoencoder-based representations such as CoMA~\cite{ranjan2018coma} define
canonical spaces for facial shape, pose, and expression modeling.
Recent topology-agnostic talking-head studies, including ScanTalk and Beyond Fixed
Topologies, further highlight the practical limitations of fixed mesh assumptions in
speech-driven 3D face animation~\cite{nocentini2024scantalk,nocentini2026beyond}.
In addition, the Procrustes/Kabsch alignment used in our normalization stage follows
standard least-squares rigid alignment formulations~\cite{kabsch1976solution,umeyama1991least}.
Rather than proposing a new registration or deformation-transfer solver, KoUniTalk
uses these classical tools to construct a compact, identity-neutral articulation space
for cross-dataset speech-driven facial animation evaluation.

\section{Method}
\label{sec:method}

\subsection{Source Datasets}

VOCASET~\cite{cudeiro2019vocaset} comprises 12 speakers across 480 sequences.
The released Korean speech-based 3D talking face data~\cite{aihub2023} provides a training set of 455,291 frames and a
validation set of 57,489 frames at 30\,fps.
Both datasets include audio corresponding to each utterance sequence, and this
correspondence is preserved after integration.
To preserve the temporal fidelity of each source, the two subsets retain their
original frame rates (60\,fps for VOCASET, 30\,fps for the released Korean speech-based data) within the unified
dataset.

\subsection{Unified Template}

The unified template is a lightweight custom 3D face mesh with 1,176 vertices and
2,236 triangles.
It intentionally focuses on the mouth and adjacent lower/mid-face region, especially
the lips and surrounding mouth geometry, while retaining nasal and mid-face context
that provides stable spatial cues for registration and rendering.
Marker vertex correspondences between each source dataset's neutral mesh and the
unified template were manually specified using Blender.

\begin{figure}[t]
  \centering
  \includegraphics[width=1\linewidth]{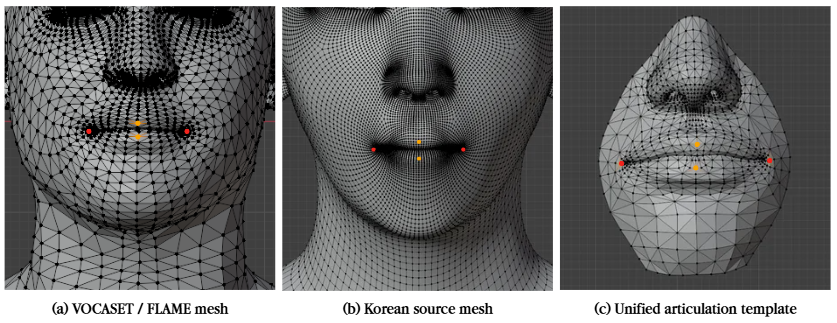}
  \caption{Source mesh and unified-template comparison. (a) VOCASET/FLAME mesh,
           (b) released Korean source mesh, and (c) the proposed unified articulation
           template. Red markers indicate mouth corners and orange markers indicate
           upper/lower lip centers used for semantic mouth-motion evaluation.}
  \label{fig:mesh_landmarks}
\end{figure}

Figure~\ref{fig:mesh_landmarks} visualizes the topology gap between the two source
meshes and the proposed unified articulation template. The red markers denote the
left and right mouth corners, while the orange markers denote the upper and lower lip
centers used for the mouth-motion preservation protocol in Sec.~\ref{sec:mouth_protocol}.

\paragraph{Scope of the unified template.}
The unified template is not intended to reproduce the full head or preserve
subject-specific identity geometry. Instead, it serves as an identity-neutral canonical
articulation space. We intentionally retain the mouth and adjacent mid-face/nasal
region, which are most directly relevant to speech-driven facial articulation and
provide stable geometric context for registration. Upper-face details such as eyes,
gaze, hair, and head shape are excluded from the target topology because they are
weakly constrained by speech and would substantially increase the output dimensionality.
Subject labels are still retained for subject-disjoint training and evaluation, but
the output geometry itself should be interpreted as canonical articulatory motion
rather than a subject-specific avatar mesh.

\subsection{Lightweight Common Representation}

The unified template is designed to reduce the output dimensionality of speech-driven
3D facial animation models while retaining the lower/mid-face geometry most relevant
to articulatory motion.
For a vertex-based regression model, the output dimensionality scales linearly with
$3V$, where $V$ is the number of vertices.
As summarized in Table~\ref{tab:efficiency}, the proposed template reduces the output
space from 15,069 dimensions in VOCASET/FLAME and 72,147 dimensions in the original
Korean mesh to 3,528 dimensions.
This dimensionality reduction is particularly useful for controlled Korean-English
training and evaluation, where both datasets must share the same output space.

\begin{table}[t]
  \centering
  \caption{Mesh and output-dimensionality comparison. Output dimension is computed as
           $3V$, where $V$ is the number of mesh vertices.}
  \label{tab:efficiency}
  \scriptsize
  \setlength{\tabcolsep}{4pt}
  \begin{adjustbox}{max width=\textwidth}
  \begin{tabular}{lrrrr}
    \toprule
    \textbf{Representation} & \textbf{Vertices} & \textbf{Faces} &
    \textbf{Output dim.} & \textbf{Relative dim.} \\
    \midrule
    VOCASET / FLAME & 5,023 & 9,976  & 15,069 & $4.27\times$ \\
    Korean original    & 24,049 & 24,002 & 72,147 & $20.45\times$ \\
    Unified (Ours)  & 1,176 & 2,236  & 3,528  & $1.00\times$ \\
    \bottomrule
  \end{tabular}
  \end{adjustbox}
\end{table}

\subsection{Deformation Transfer}

The objective function $E$ is defined as a weighted sum of three energy terms.
The smoothness energy $E_s$ encourages smooth variation of deformation between
adjacent triangles by minimizing the change in deformation rather than the mesh
geometry itself.
The identity-transformation energy $E_I$ keeps the triangle transformations in
unconstrained regions close to the identity matrix, preventing excessive shape
distortion induced by $E_s$.
The closest-valid-point energy $E_c$ drives each vertex of the source mesh toward
the closest valid point on the target mesh.
Here, $w_S$, $w_I$, and $w_C$ denote the respective weights.
Because the Korean source mesh is represented with quadrilateral faces whereas
Deformation Transfer is defined over triangular elements, we triangulate the Korean
connectivity using a deterministic diagonal split before correspondence search and
per-frame affine transformation estimation. The resulting triangulated connectivity is
kept fixed for all Korean frames.
Deformation Transfer~\cite{sumner2004deformation} is applied to each source dataset
in the following three steps.

\begin{enumerate}
  \item \textbf{Correspondence establishment.}
        Minimize
        $E = w_S \cdot E_s + w_I \cdot E_I + w_C \cdot E_c$
        in two stages.
        In stage~1, $w_C = 0$ so that only marker constraints are used;
        in stage~2, $w_C$ is gradually increased for precise local registration.
        This step is performed once per source--target pair.

  \item \textbf{Triangle correspondence search.}
        Correspondence pairs between the deformed source and the unified template are
        established based on triangle centroid distance and normal direction, and are
        reused across all frames together with the stored LU factorization.

  \item \textbf{Deformation transfer.}
        Per-frame affine transformations for each triangle are computed from the
        correspondences, and deformed vertex positions on the unified template are
        obtained via LU back-substitution using the stored factorization.
\end{enumerate}

\subsection{Parameter Design for Correspondence Establishment}

Parameters were set independently for VOCASET and the released Korean speech-based data according to the vertex density
and geometric complexity of each source mesh.
Table~\ref{tab:params} summarizes the applied parameters.

\begin{table}[t]
  \centering
  \caption{Parameters for the correspondence establishment stage.}
  \label{tab:params}
  \begin{adjustbox}{max width=\textwidth}
  \begin{tabular}{ll}
    \toprule
    \textbf{Parameter} & \textbf{Value} \\
    \midrule
    $w_S$  & $5\times10^{-6}$ -- $1\times10^{-4}$ \\
    $w_I$  & $1\times10^{-7}$ -- $5\times10^{-7}$ \\
    $w_C$ (4 stages) & $(1{\sim}10.0)\!\to\!50\!\to\!200.0\!\to\!500.0$ \\
    $k$    & $500$--$1{,}200$ \\
    $\mathrm{thresh}_\mathrm{dist}$ & $0.5$ \\
    \bottomrule
  \end{tabular}
  \end{adjustbox}
\end{table}

$w_C$ is progressively increased across four stages in the second phase:
$(1{\sim}10.0) \to 50 \to 200.0 \to 500.0$.
Global shape alignment is first achieved with a low initial $w_C$, after which
stepwise increases enable precise local registration, thereby avoiding convergence
instability and local minima.

Since the Korean mesh has approximately 20-fold the vertex density of the unified template,
$k$ and $\mathrm{thresh\_dist}$ were found to be sensitive to registration quality.
In contrast, VOCASET has fewer vertices, resulting in comparatively higher convergence
stability with respect to parameter variation.

\subsection{Procrustes Alignment}

After Deformation Transfer, the two resulting mesh sets---one derived from VOCASET
and the other from the released Korean speech-based data---reside in different coordinate frames due to the substantial
differences in scale and orientation between the two source datasets.
Direct concatenation without further normalization would therefore prevent a model
from learning consistent motion representations across languages.

To resolve this, we apply Procrustes Alignment to bring both mesh sets into a common
coordinate frame.
The alignment is performed in three steps with respect to the neutral mesh of each
source:

\begin{enumerate}
  \item \textbf{Translation.}
        The centroid of each neutral mesh is translated to the origin so that all
        sequences share a common reference point.

  \item \textbf{Scale normalization.}
        Each mesh is scaled by the inverse of its root-mean-square (RMS) vertex
        distance from the origin, normalizing the overall size across datasets.

  \item \textbf{Rotation (SVD).}
        The optimal rotation matrix $\mathbf{R}$ that aligns the neutral mesh to a
        reference orientation is computed via Singular Value Decomposition (SVD).
        A Y-axis flip is automatically detected and corrected during this step.
        The transformation derived from the neutral mesh is then applied uniformly to
        all frames of the corresponding sequence, preserving temporal consistency.
\end{enumerate}

This per-source normalization ensures that the VOCASET-derived and Korean-derived meshes
share the same scale and orientation in the unified dataset, enabling direct use for
multilingual model training without additional topology conversion.

\section{Experiments}
\label{sec:exp}

\subsection{Dataset Statistics Comparison}

Table~\ref{tab:stats} compares the key statistics of the two source datasets and the
unified dataset.
The two source datasets differ across most dimensions---language, speaker count,
vertex count, and FPS---and this heterogeneity is the primary obstacle to direct
integration.
In particular, VOCASET uses triangular meshes while the released Korean speech-based data uses quadrilateral meshes,
introducing structural incompatibility beyond a simple vertex count mismatch.

\begin{table}[t]
  \centering
  \caption{Dataset statistics comparison. Source inventory counts and exact processed
           matched counts are reported separately because they are defined over
           different scopes.}
  \label{tab:stats}
  \scriptsize
  \setlength{\tabcolsep}{3pt}
  \begin{adjustbox}{max width=\textwidth}
  \begin{tabular}{lccc}
    \toprule
    & \textbf{VOCASET} & \makecell{\textbf{Korean speech-based}\\\textbf{data}} & \textbf{Unified (Ours)} \\
    \midrule
    Language         & English  & Korean   & English + Korean \\
    Speakers         & 12       & 10       & 22 \\
    Source sequences & 480      & 4,500$^\dagger$  & 4,980 \\
    Source-reported frames & $\sim$104,400$^\ddagger$ & 512,780$^{\dagger,\ddagger}$ & $\sim$617,180 \\
    Processed matched sequences & 478 & 4,500 & 4,978 \\
    Processed matched frames & 123,341 & 519,440 & 642,781 \\
    Vertices         & 5,023    & 24,049   & 1,176 (Ours) \\
    Faces            & 9,976    & 24,002$^\S$   & 2,236 \\
    FPS              & 60       & 30       & 60 / 30 (preserved) \\
    Topology         & FLAME    & own template & Ours \\
    Released split   & full     & train+val & -- \\
    \bottomrule
  \end{tabular}
  \end{adjustbox}

  \smallskip
  {\footnotesize
  $^\dagger$ Only the Korean train+validation sets are used (test set not publicly released).\\
  $^\ddagger$ Source-reported frame counts follow the released metadata or the
  commonly reported approximate VOCASET count. Processed counts are exact counts of
  frame-matched original/retargeted pairs enumerated by our pipeline and used in the
  preservation audit; they are not obtained by summing the source-reported estimates.\\
  $^\S$ Korean face count denotes the original quadrilateral connectivity; the mesh is triangulated internally for deformation transfer.}
\end{table}

The processed matched inventory contains 22 speakers, 4,978 sequences, and 642,781
frames. Of these, 478 sequences and 123,341 frames are from VOCASET, and 4,500
sequences and 519,440 frames are from the Korean subset. The approximately 617,180
frames above instead summarize the two source-reported inventories and are retained
only for source-level comparison.
All meshes are aligned to the unified template and can be used for model training
without additional topology conversion. Speaker ID and language labels are retained
for subject-disjoint splits and cross-domain analysis, while the mesh geometry itself
is represented in an identity-neutral articulation space.

\subsection{Benchmark Splits}
\label{sec:splits}

For VOCASET, we follow the commonly used FaceFormer subject-disjoint split.
For the released Korean speech-based data, the official test set is not publicly released; therefore, we define a new
subject-disjoint split on the released Korean subset.
This protocol is not intended to reproduce the official evaluation protocol for the released Korean speech-based data, but to provide
a controlled Korean test set compatible with the VOCASET subject-disjoint evaluation
setting.
We use speakers M01--M08 for training, M09 for validation, and M10 for testing; folders
with the ``\_val'' suffix are assigned to the same split as their corresponding speaker.
Table~\ref{tab:splits} summarizes the benchmark split used in this work.

\begin{table}[t]
  \centering
  \caption{Subject-disjoint benchmark split. For the released Korean speech-based data, folders with the ``\_val'' suffix
           are assigned to the same split as their corresponding subject.}
  \label{tab:splits}
  \scriptsize
  \setlength{\tabcolsep}{4pt}
  \begin{adjustbox}{max width=\textwidth}
  \begin{tabular}{lllll}
    \toprule
    \textbf{Dataset} & \textbf{Protocol} & \textbf{Train} & \textbf{Val} & \textbf{Test} \\
    \midrule
    VOCASET & FaceFormer split & 8 FaceTalk subjects & 2 FaceTalk subjects & 2 FaceTalk subjects \\
    Korean speech-based data & Proposed split & M01--M08 & M09 & M10 \\
    \bottomrule
  \end{tabular}
  \end{adjustbox}
\end{table}

\paragraph{Availability.}
Upon acceptance, we will release the unified articulation template, split files,
evaluation scripts, and preprocessing code. Redistribution of source-derived mesh
sequences follows the license terms of the original datasets; when direct redistribution
is restricted, the released code will reproduce the retargeted sequences from the
corresponding licensed source data.

\subsection{Qualitative Retargeting Results}
\label{sec:qualitative_results}

\begin{figure}[ht]
  \centering
  \includegraphics[width=0.95\linewidth]{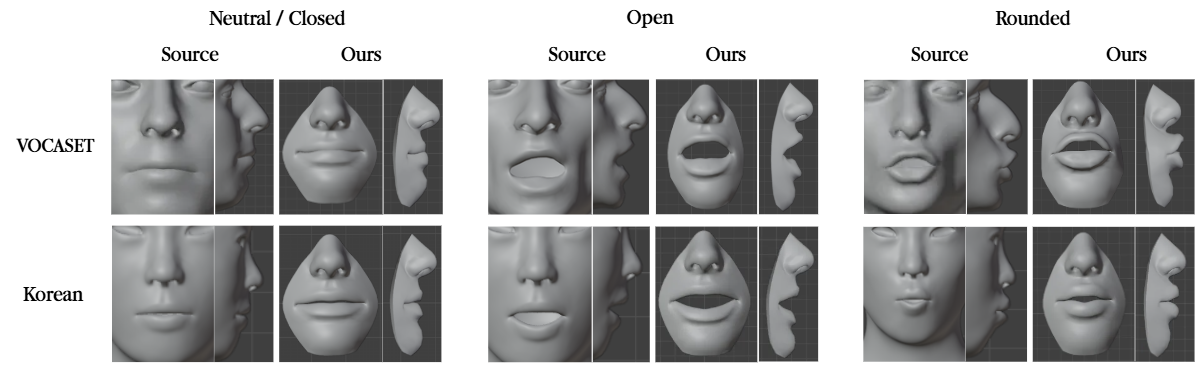}
  \caption{Qualitative retargeting examples under different mouth configurations.
           For each dataset, we show the original source mesh and the corresponding
           retargeted result on our unified articulation template. The examples include
           neutral/closed, open, and rounded mouth configurations, illustrating that
           visible articulation patterns from both VOCASET and Korean source meshes are
           transferred to the lightweight template.}
  \label{fig:result}
\end{figure}

Fig.~\ref{fig:result} shows qualitative retargeting examples for both VOCASET and the
Korean source data under neutral/closed, open, and rounded mouth configurations. The
examples illustrate that visible speech-related mouth configurations from the original
source meshes are transferred to the lightweight unified articulation template.

\subsection{Mouth Motion Preservation Protocol}
\label{sec:mouth_protocol}

Since the source and retargeted meshes have different topologies, we do not compute
dense vertex-wise errors. Instead, we evaluate whether speech-related articulation is
preserved using semantic mouth-landmark trajectories. This landmark-based design is
motivated by lip-sync, viseme, and evaluation studies that emphasize visible mouth
shapes and audio-visual synchrony rather than pointwise vertex identity~\cite{edwards2016jali,chung2017outoftime,prajwal2020wav2lip,haque2025wildwest}.
For each sequence, we extract the two mouth corners and the upper/lower lip-center
vertices, and compute mouth width, mouth opening, and aperture ratio over time.
The landmark indices used in our experiments are listed in Table~\ref{tab:landmarks}.

\begin{table}[H]
  \centering
  \caption{Semantic mouth landmarks used for motion-preservation evaluation.}
  \label{tab:landmarks}
  \scriptsize
  \begin{adjustbox}{max width=\textwidth}
  \begin{tabular}{lrrrr}
    \toprule
    \textbf{Topology} & \textbf{Left corner} & \textbf{Right corner} &
    \textbf{Upper lip} & \textbf{Lower lip} \\
    \midrule
    VOCASET original & 487 & 198 & 688 & 667 \\
    Korean original     & 4707 & 3010 & 4118 & 910 \\
    Unified (Ours)   & 327 & 467 & 864 & 151 \\
    \bottomrule
  \end{tabular}
  \end{adjustbox}
\end{table}

For each sequence, we compute mouth opening, mouth width, aperture ratio, and opening
velocity as
\begin{gather}
d_{\mathrm{open}}(t)
= \left\|\mathbf{v}_{\mathrm{upper}}(t)
- \mathbf{v}_{\mathrm{lower}}(t)\right\|_2,
\label{eq:mouth_open} \\
d_{\mathrm{width}}(t)
= \left\|\mathbf{v}_{\mathrm{left}}(t)
- \mathbf{v}_{\mathrm{right}}(t)\right\|_2,
\label{eq:mouth_width} \\
r_{\mathrm{aperture}}(t)
= \frac{d_{\mathrm{open}}(t)}{d_{\mathrm{width}}(t)},
\label{eq:mouth_ratio} \\
v_{\mathrm{open}}(t)
= d_{\mathrm{open}}(t) - d_{\mathrm{open}}(t-1).
\label{eq:mouth_velocity}
\end{gather}
Opening and width trajectories are normalized by the corresponding first-frame mouth
width, and the aperture-ratio trajectory is centered by subtracting its first-frame
value. We then compare the original and retargeted trajectories using Pearson
correlation, z-normalized RMSE, amplitude error, and velocity correlation computed on
first-order temporal differences.

For this audit, we use all original/retargeted pairs available in the processed
matched inventory: 478 VOCASET sequences (123,341 frames) and 4,500 Korean sequences
(519,440 frames). All included pairs have matched frame counts, enabling direct
frame-aligned comparison of semantic landmark trajectories. We aggregate sequence
metrics within each subject and report the mean and standard deviation across
subject-level summaries.

\begin{table}[H]
  \centering
  \caption{Full-set mouth-motion preservation audit between original and retargeted
           sequences. All available processed matched pairs are compared using
           semantic mouth-landmark curves. Values are mean$\pm$std across
           subject-level summaries.}
  \label{tab:mouth_preservation}
  \scriptsize
  \setlength{\tabcolsep}{3pt}
  \begin{adjustbox}{max width=\textwidth}
  \begin{tabular}{lllrrrcccc}
    \toprule
    \textbf{Dataset} & \textbf{Transfer} & \textbf{Curve} & \textbf{\#Subj.} &
    \textbf{\#Seq.} & \textbf{\#Frames} &
    \textbf{Corr.} $\uparrow$ & \textbf{z-RMSE} $\downarrow$ &
    \textbf{Amp. Err.} $\downarrow$ & \textbf{Vel. Corr.} $\uparrow$ \\
    \midrule
    VOCASET & original $\rightarrow$ ours & Opening & 12 & 478 & 123,341 & $0.9890{\pm}0.0113$ & $0.1302{\pm}0.0593$ & $0.2239{\pm}0.1101$ & $0.9864{\pm}0.0124$ \\
    VOCASET & original $\rightarrow$ ours & Width & 12 & 478 & 123,341 & $0.6489{\pm}0.1245$ & $0.8044{\pm}0.1517$ & $0.4877{\pm}0.1961$ & $0.5995{\pm}0.1139$ \\
    VOCASET & original $\rightarrow$ ours & Aperture ratio & 12 & 478 & 123,341 & $0.9899{\pm}0.0073$ & $0.1289{\pm}0.0473$ & $0.2064{\pm}0.0861$ & $0.9889{\pm}0.0096$ \\
    Korean & original $\rightarrow$ ours & Opening & 10 & 4,500 & 519,440 & $0.9929{\pm}0.0067$ & $0.1018{\pm}0.0508$ & $0.5682{\pm}0.0208$ & $0.9957{\pm}0.0034$ \\
    Korean & original $\rightarrow$ ours & Width & 10 & 4,500 & 519,440 & $0.8871{\pm}0.0767$ & $0.4332{\pm}0.1512$ & $0.4514{\pm}0.0788$ & $0.8255{\pm}0.0749$ \\
    Korean & original $\rightarrow$ ours & Aperture ratio & 10 & 4,500 & 519,440 & $0.9915{\pm}0.0055$ & $0.1210{\pm}0.0377$ & $0.2819{\pm}0.0515$ & $0.9928{\pm}0.0034$ \\
    \bottomrule
  \end{tabular}
  \end{adjustbox}
\end{table}

Table~\ref{tab:mouth_preservation} shows strong preservation of the temporal shape and
timing of vertical mouth opening and aperture ratio. Amplitude is not uniformly
preserved, especially for Korean opening, and mouth width is substantially weaker,
particularly for VOCASET. We therefore treat width as a diagnostic sensitive to
lateral scale and mouth-corner correspondence rather than claiming uniform
preservation of every articulation component.

\subsection{Downstream Compatibility Evaluation}
\label{sec:downstream_protocol}

The unified topology enables speech-driven 3D facial animation models to be trained
and evaluated under a single Korean-English output space.
Since the goal of KoUniTalk is not to propose a new generator architecture, the
downstream experiment is designed as a compatibility evaluation for the proposed
articulation-template space.

We evaluate an audio-agnostic fixed-motion control together with SelfTalk and
CodeTalker as representative speech-driven 3D facial animation baselines adapted to
the proposed 3,528-dimensional unified-template output space.
For CodeTalker, we report the primary all-condition evaluation rather than the
best-condition-per-utterance variant to avoid cherry-picking across generated
conditions. For both learning-based baselines, the dataset-specific output layer is
replaced with a 3,528-dimensional regression head corresponding to the 1,176-vertex
unified template, and all configurations are trained and evaluated in the same
Procrustes-aligned coordinate frame using the subject-disjoint splits defined in
Sec.~\ref{sec:splits}.

We define three training configurations: VOCASET-only, Korean-only, and pooled
VOCASET+Korean training.
In the mixed setting, the two training subsets are concatenated and shuffled without
domain-balanced sampling.
Consequently, the mixed model is trained with the natural dataset proportions, where
the Korean subset is substantially larger than VOCASET.
We therefore interpret this setting as a Korean-dominant pooled mixed baseline rather
than a balanced bilingual training result.
The audio--mesh-paired training inventory used by the learning-based baseline contains
314 VOCASET and 3,200 Korean sequences. These counts are smaller than the processed
mesh inventory in Table~\ref{tab:stats} because model training additionally requires a
valid paired audio record.

Evaluation is reported separately on the English VOCASET test split and the Korean
test split.
This train/evaluation matrix separates three effects: in-domain performance
(VOCASET-only on English and Korean-only on Korean), cross-domain generalization
(VOCASET-only on Korean and Korean-only on English), and pooled mixed training
(VOCASET+Korean on both test domains).
We report all-vertex and lip-region coordinate mean-squared errors over the unified
template, denoted as MVE and LVE, respectively.
Following the result sheet, all LVE/MVE values in Table~\ref{tab:downstream} are
multiplied by $10^{6}$ for readability.
Macro LVE is the arithmetic mean of English and Korean LVE, and Worst LVE is the
larger of the two.

\begin{table}[t]
  \centering
  \caption{Downstream compatibility evaluation on English and Korean test sets.
           All predictions and ground truths are represented in the unified
           ours-template topology. LVE/MVE values are coordinate MSE $\times 10^{6}$
           over the lip region and all vertices, respectively. Macro and Worst are
           computed from English/Korean LVE. The VOCASET+Korean setting uses
           natural-ratio pooled training, not balanced 1:1 sampling.}
  \label{tab:downstream}
  \scriptsize
  \setlength{\tabcolsep}{4pt}
  \begin{adjustbox}{max width=\textwidth}
  \begin{tabular}{lllrrrrrr}
    \toprule
    \textbf{Model} & \textbf{Train data} & \textbf{Sampling} &
    \textbf{En LVE} $\downarrow$ & \textbf{Ko LVE} $\downarrow$ &
    \textbf{Macro LVE} $\downarrow$ & \textbf{Worst LVE} $\downarrow$ &
    \textbf{En MVE} $\downarrow$ & \textbf{Ko MVE} $\downarrow$ \\
    \midrule
    Neutral template & Mean motion & -- & 4.19 & 13.58 & 8.88 & 13.58 & 2.57 & 13.23 \\
    SelfTalk & VOCASET only & -- & 1.99 & 14.00 & 7.99 & 14.00 & 1.34 & 13.54 \\
    SelfTalk & Korean only & -- & 9.45 & 1.22 & 5.33 & 9.45 & 9.32 & 1.82 \\
    SelfTalk & VOCASET+Korean & natural ratio & 2.46 & 2.30 & 2.38 & 2.46 & 1.61 & 2.89 \\
    CodeTalker & VOCASET only & -- & 2.62 & 31.07 & 16.85 & 31.07 & 1.76 & 29.87 \\
    CodeTalker & Korean only & -- & 7.40 & 7.22 & 7.31 & 7.40 & 6.95 & 7.54 \\
    CodeTalker & VOCASET+Korean & natural ratio & 2.43 & 4.82 & 3.62 & 4.82 & 1.68 & 5.21 \\
    \bottomrule
  \end{tabular}
  \end{adjustbox}
\end{table}

The single-source rows reveal substantial cross-domain degradation: for example,
SelfTalk trained on VOCASET obtains 1.99 LVE on the English test split but 14.00 LVE
on the Korean test split, while SelfTalk trained on Korean obtains 1.22 LVE on Korean
but 9.45 LVE on English.
CodeTalker shows a similar source sensitivity, with 2.62/31.07 LVE for
VOCASET-only training on English/Korean and 7.40/7.22 LVE for Korean-only training.
The natural-ratio pooled setting reduces worst-domain errors compared with these
single-source cross-domain baselines, but it is not necessarily the per-domain optimum
for every model.
This supports our interpretation that KoUniTalk enables direct Korean-English
training and evaluation in a shared topology, while the choice of sampling strategy
remains important for optimizing per-domain accuracy.

To support this interpretation, we analyze the ground-truth motion distributions over
the retargeted sequences for which the motion statistics are available. This
motion-analysis inventory is distinct from the audio--mesh-paired model-training
inventory described above.
As shown in Table~\ref{tab:gt_motion_distribution}, the Korean training split
contains 3,599 sequences, whereas the VOCASET training split contains 319 sequences,
corresponding to an 11.28$\times$ sequence imbalance.
The two domains also differ in native-frame motion scale: in the training split, the
Korean subset has 1.40$\times$ larger mean lip displacement and 2.00$\times$ larger
mean mouth-opening displacement than VOCASET.
However, this relationship is not fixed across splits; in the test split, VOCASET has
1.88$\times$ larger mean lip displacement than the Korean test split.
Because the two source frame rates are preserved, these statistics should be interpreted
as native-frame displacement measurements rather than physical per-second velocities.
They indicate that the unified topology removes mesh-format incompatibility, but it does
not remove domain imbalance or motion-distribution mismatch.

\begin{table}[t]
  \centering
  \caption{Ground-truth mouth-motion distribution after retargeting to the unified
           topology. Values are computed on the retargeted sequences. Lip displacement
           denotes the mean native-frame lip displacement magnitude. Because VOCASET and
           Korean sequences retain different frame rates, these values are not physical
           per-second velocities. The natural training set is highly imbalanced toward
           the Korean subset, and the relative motion scale differs across train/test
           splits.}
  \label{tab:gt_motion_distribution}
  \scriptsize
  \setlength{\tabcolsep}{4pt}
  \begin{adjustbox}{max width=\textwidth}
  \begin{tabular}{llrrrrr}
    \toprule
    \textbf{Dataset} & \textbf{Split} & \textbf{Seq.} &
    \textbf{Lip disp.} $\uparrow$ & \textbf{Open disp.} $\uparrow$ &
    \textbf{Mouth open} $\uparrow$ & \textbf{Aperture} $\uparrow$ \\
    \midrule
    VOCASET & train & 319 & 0.000500 & 0.000702 & 0.0126 & 0.2241 \\
    Korean & train & 3,599 & 0.000700 & 0.001402 & 0.0155 & 0.2848 \\
    VOCASET & test & 79 & 0.000776 & 0.001348 & 0.0164 & 0.2966 \\
    Korean & test & 450 & 0.000413 & 0.001021 & 0.0134 & 0.2338 \\
    \bottomrule
  \end{tabular}
  \end{adjustbox}
\end{table}

For model outputs, we further compute a predicted-to-ground-truth motion ratio from
lip displacement magnitudes.
A ratio below 1 indicates under-articulation or overly static predictions, whereas a
ratio above 1 indicates exaggerated motion.
As shown in Table~\ref{tab:motion_ratio}, the pooled mixed models exhibit
source-dependent motion-scale bias: SelfTalk under-drives VOCASET motion
(0.62) while over-driving Korean motion (1.68), and CodeTalker shows the same trend
(0.69 on VOCASET and 1.60 on Korean).
This explains why natural-ratio mixed training does not uniformly improve LVE/MVE
over domain-specific training, even though it enables a single model to be evaluated
on both domains.

\begin{table}[t]
  \centering
  \caption{Predicted-to-ground-truth lip motion ratio on English and Korean test
           sets. Ratios below 1 indicate under-articulation or overly static motion,
           while ratios above 1 indicate over-driving or noisy motion scale. The
           VOCASET+Korean setting uses natural-ratio pooled training.}
  \label{tab:motion_ratio}
  \scriptsize
  \setlength{\tabcolsep}{6pt}
  \begin{adjustbox}{max width=\textwidth}
  \begin{tabular}{lllrr}
    \toprule
    \textbf{Model} & \textbf{Train data} & \textbf{Sampling} &
    \textbf{En ratio} $\rightarrow 1$ & \textbf{Ko ratio} $\rightarrow 1$ \\
    \midrule
    Neutral template & Mean motion & -- & 0.00 & 0.00 \\
    SelfTalk & VOCASET only & -- & 0.74 & 1.21 \\
    SelfTalk & Korean only & -- & 1.65 & 1.33 \\
    SelfTalk & VOCASET+Korean & natural ratio & 0.62 & 1.68 \\
    CodeTalker & VOCASET only & -- & 0.69 & 1.74 \\
    CodeTalker & Korean only & -- & 0.58 & 1.46 \\
    CodeTalker & VOCASET+Korean & natural ratio & 0.69 & 1.60 \\
    \bottomrule
  \end{tabular}
  \end{adjustbox}
\end{table}

Taken together, Tables~\ref{tab:downstream}--\ref{tab:motion_ratio} should be
interpreted as a diagnostic compatibility evaluation rather than a claim that pooled
training is the optimal bilingual training strategy.
The natural-ratio union enables a single model to be trained and evaluated across
English and Korean in one compact articulation space, but balanced or language-aware
sampling remains necessary for optimizing per-domain accuracy.
This is a limitation of the current downstream baseline, not a failure of the
topology integration itself.

\section{Limitations}
\label{sec:limitations}

KoUniTalk uses a single lightweight canonical template, which improves topology
compatibility and reduces output dimensionality but does not preserve subject-specific
full-head geometry, eye motion, gaze, or upper-face expressions.
The benchmark should therefore be interpreted as an articulation-centered evaluation
space rather than a complete identity-preserving avatar representation.
Subject labels are retained for subject-disjoint splitting and cross-speaker analysis,
but the output mesh geometry is identity-neutral.
Extending KoUniTalk to subject-specific neutral templates, such as FLAME-topology
identities with neutral-conditioned generation, is an important direction for future
work.

The template retains the lips and adjacent lower/mid-face context but omits broad
lateral cheeks and the upper face. Its 1,176-vertex resolution was not selected by a
resolution ablation, and the present landmark protocol does not measure
anterior--posterior lip protrusion. Accordingly, the current results do not establish
fidelity for strongly rounded or puckered mouth shapes. The four-landmark protocol is
also sparse and should be read as a trajectory diagnostic rather than a dense
mouth-surface validation.

Our downstream evaluation is not intended to be exhaustive. Because existing
speech-driven 3D facial animation models often rely on dataset-specific output
topologies, adapting every architecture to a newly unified template requires
non-trivial engineering and computational cost. We therefore report an initial set of
representative baselines together with diagnostic motion-scale analyses, and leave a broader model zoo
for future extensions of KoUniTalk. KoUniTalk removes mesh-topology mismatch, but it
does not remove linguistic, recording, or motion-distribution bias between the two
source datasets.
Moreover, language co-varies with speaker identity, capture setup, frame rate, scale,
and motion distribution in the two source datasets. The observed differences are
therefore cross-dataset/domain effects and cannot be interpreted as causal effects of
language. The Korean validation and test splits each contain only one speaker, so the
reported results do not quantify variation across alternative speaker splits.

\paragraph{Ethical considerations.}
We use publicly released speech-driven 3D facial datasets and do not collect new
human-subject data. Source data are processed according to their respective dataset
terms. The proposed benchmark is intended for research on speech-driven facial
articulation, and the unified template is identity-neutral rather than designed to
reconstruct or preserve subject-specific full-head identity.

\section{Conclusion}
\label{sec:conclusion}

We presented KoUniTalk, a lightweight articulation-centered Korean-English 3D talking
face benchmark constructed by retargeting VOCASET and the released Korean speech-based 3D talking face data onto a single unified topology using Deformation Transfer and Procrustes
Alignment.
Rather than proposing a new deformation transfer solver, this work focuses on building
an identity-neutral common representation for controlled training and evaluation
across datasets that originally have incompatible mesh structures.
The unified representation reduces the output dimensionality from 15,069 dimensions in
VOCASET/FLAME and 72,147 dimensions in the original Korean mesh to 3,528 dimensions,
while retaining the mouth and adjacent lower/mid-face geometry required for
speech-driven articulation analysis.
We also define a subject-disjoint Korean evaluation split for the released Korean subset
and a semantic mouth-landmark protocol for evaluating motion preservation after
retargeting.

The downstream compatibility protocol further provides an initial basis for evaluating
future speech-driven 3D facial animation models under a shared Korean-English
articulation space.


\bibliographystyle{splncs04}
\bibliography{main}

\clearpage
\appendix
\setcounter{section}{0}
\setcounter{figure}{0}
\setcounter{table}{0}
\setcounter{equation}{0}
\renewcommand{\thesection}{S\arabic{section}}
\renewcommand{\thefigure}{S\arabic{figure}}
\renewcommand{\thetable}{S\arabic{table}}
\renewcommand{\theequation}{S\arabic{equation}}
\section*{Supplementary Material}
\addcontentsline{toc}{section}{Supplementary Material}

\section{Overview}
\label{sec:supp_overview}

This supplementary material supports the main submission with additional details and
analysis for the KoUniTalk benchmark. The main paper is self-contained; this document
provides additional evidence for retargeting quality and reproducibility rather than a
revised method or a new central claim.

\section{Unified Template, Landmarks, and Lip Mask}
\label{sec:supp_template}

KoUniTalk uses a lightweight unified articulation template with 1,176 vertices and
2,236 triangular faces. The resulting output dimensionality is $1{,}176\times3=3{,}528$.
The template is intended as an identity-neutral canonical output space for
speech-related mouth and lower/mid-face articulation.

\begin{table}[H]
  \centering
  \caption{Semantic mouth landmarks used to evaluate mouth-motion preservation.
  The original and retargeted meshes have different topologies, so preservation is
  measured through these semantic landmark trajectories rather than through direct
  vertex-wise comparison.}
  \label{tab:supp_landmarks}
  \scriptsize
  \begin{adjustbox}{max width=\textwidth}
  \begin{tabular}{lrrrr}
    \toprule
    \textbf{Topology} & \textbf{Left corner} & \textbf{Right corner} &
    \textbf{Upper lip} & \textbf{Lower lip} \\
    \midrule
    VOCASET original & 487 & 198 & 688 & 667 \\
    Korean original & 4707 & 3010 & 4118 & 910 \\
    Unified template & 327 & 467 & 864 & 151 \\
    \bottomrule
  \end{tabular}
  \end{adjustbox}
\end{table}

\paragraph{Lip-region mask.}
For downstream lip-region evaluation, let $\mathcal{L}$ denote the fixed lip-region
vertex set on the unified template. The mask contains 262 vertices and was manually
selected on the unified template to include the lip contour, inner mouth boundary, and
immediate perioral vertices used for lip-region evaluation. The indices are 0-based,
range from 27 to 1,175, and contain no duplicates. The same mask is used for all
datasets and baselines when computing lip-region vertex error (LVE). The 0-based
vertex indices and metadata are provided in
\nolinkurl{anc/code/splits/lip_mask_indices.txt} and
\nolinkurl{anc/code/splits/lip_mask_metadata.json}, respectively.

\section{Benchmark Split and Dataset Statistics}
\label{sec:supp_splits}

For VOCASET, we follow the subject-disjoint split used in prior speech-driven 3D facial
animation work. For the released Korean speech-based 3D facial motion data, the
official test set is not publicly released; therefore, KoUniTalk defines a
subject-disjoint split from the released data. The Korean split uses M01--M08 for
training, M09 for validation, and M10 for testing. Folders with the ``\_val'' suffix are
assigned to the same split as their corresponding speaker.

\begin{table}[H]
  \centering
  \caption{Benchmark split statistics computed from the processed split records.
  Duration is reported in minutes.}
  \label{tab:supp_splits}
  \scriptsize
  \setlength{\tabcolsep}{4pt}
  \begin{adjustbox}{max width=\textwidth}
  \begin{tabular}{lllrrrr}
    \toprule
    \textbf{Dataset} & \textbf{Split} & \textbf{Subjects} & \textbf{FPS} &
    \textbf{Seq.} & \textbf{Frames} & \textbf{Duration (min)} \\
    \midrule
    VOCASET & Train & 8 speakers & 60 & 319 & 85,064 & 23.6 \\
    VOCASET & Val & 2 speakers & 60 & 80 & 20,079 & 5.6 \\
    VOCASET & Test & 2 speakers & 60 & 79 & 18,198 & 5.1 \\
    Korean & Train & M01--M08 & 30 & 3,600 & 459,645 & 255.4 \\
    Korean & Val & M09 & 30 & 450 & 32,387 & 18.0 \\
    Korean & Test & M10 & 30 & 450 & 27,408 & 15.2 \\
    \midrule
    VOCASET & Total & 12 speakers & 60 & 478 & 123,341 & 34.3 \\
    Korean & Total & 10 speakers & 30 & 4,500 & 519,440 & 288.6 \\
    Combined & Total & 22 speakers & 60 / 30 & 4,978 & 642,781 & 322.9 \\
    \bottomrule
  \end{tabular}
  \end{adjustbox}
\end{table}

\section{Extended Mouth-Motion Preservation Evaluation}
\label{sec:supp_mouth}

The main paper reports the full-set mouth-motion preservation audit over the processed
matched inventory: 478 VOCASET sequences from 12 subjects and 4,500 Korean sequences
from 10 subjects. We reproduce the detailed results here alongside the supplementary
protocol and visualizations. The exact counts are 123,341 VOCASET frames and 519,440
Korean frames, for a combined total of 4,978 sequences and 642,781 frames.

Because the source meshes and unified-template meshes have different topologies, we
compare semantic mouth-motion curves. For each sequence, we compute
\begin{gather}
  d_{\mathrm{open}}(t)=\|\mathbf{v}_{\mathrm{upper}}(t)-\mathbf{v}_{\mathrm{lower}}(t)\|_2, \\
  d_{\mathrm{width}}(t)=\|\mathbf{v}_{\mathrm{left}}(t)-\mathbf{v}_{\mathrm{right}}(t)\|_2, \\
  r_{\mathrm{aperture}}(t)=\frac{d_{\mathrm{open}}(t)}{d_{\mathrm{width}}(t)}.
\end{gather}
For each signal $s(t)\in\{d_{\mathrm{open}}(t), d_{\mathrm{width}}(t),
r_{\mathrm{aperture}}(t)\}$, we also compute the temporal difference
\begin{equation}
  \Delta s(t)=s(t)-s(t-1),
\end{equation}
and report the correlation between original and retargeted temporal-difference curves
as velocity correlation.

\begin{table}[H]
  \centering
  \caption{Extended mouth-motion preservation results. Values are reported as
  mean$\pm$std across subject-level summaries.}
  \label{tab:supp_mouth_preservation}
  \scriptsize
  \setlength{\tabcolsep}{2.4pt}
  \begin{adjustbox}{max width=\textwidth}
  \begin{tabular}{lllrrrcccc}
    \toprule
    \textbf{Dataset} & \textbf{Sampling} & \textbf{Signal} & \textbf{\#Subj.} &
    \textbf{\#Seq.} & \textbf{\#Frames} & \textbf{Corr.} $\uparrow$ &
    \textbf{z-RMSE} $\downarrow$ & \textbf{Amp. Err.} $\downarrow$ &
    \textbf{Vel. Corr.} $\uparrow$ \\
    \midrule
    VOCASET & Matched set & Opening & 12 & 478 & 123,341 & $0.9890\!\pm\!0.0113$ & $0.1302\!\pm\!0.0593$ & $0.2239\!\pm\!0.1101$ & $0.9864\!\pm\!0.0124$ \\
    VOCASET & Matched set & Width & 12 & 478 & 123,341 & $0.6489\!\pm\!0.1245$ & $0.8044\!\pm\!0.1517$ & $0.4877\!\pm\!0.1961$ & $0.5995\!\pm\!0.1139$ \\
    VOCASET & Matched set & Aperture & 12 & 478 & 123,341 & $0.9899\!\pm\!0.0073$ & $0.1289\!\pm\!0.0473$ & $0.2064\!\pm\!0.0861$ & $0.9889\!\pm\!0.0096$ \\
    Korean & Matched set & Opening & 10 & 4,500 & 519,440 & $0.9929\!\pm\!0.0067$ & $0.1018\!\pm\!0.0508$ & $0.5682\!\pm\!0.0208$ & $0.9957\!\pm\!0.0034$ \\
    Korean & Matched set & Width & 10 & 4,500 & 519,440 & $0.8871\!\pm\!0.0767$ & $0.4332\!\pm\!0.1512$ & $0.4514\!\pm\!0.0788$ & $0.8255\!\pm\!0.0749$ \\
    Korean & Matched set & Aperture & 10 & 4,500 & 519,440 & $0.9915\!\pm\!0.0055$ & $0.1210\!\pm\!0.0377$ & $0.2819\!\pm\!0.0515$ & $0.9928\!\pm\!0.0034$ \\
    \bottomrule
  \end{tabular}
  \end{adjustbox}
\end{table}

The full-set audit shows strong preservation of the temporal shape and timing of
vertical mouth opening and aperture-ratio dynamics on both source datasets. Amplitude
is not uniformly preserved, especially for Korean opening. The mouth-width signal is
reported as a diagnostic because it is more sensitive to lateral scale and landmark
placement across the source and unified topologies; this sensitivity is especially
visible in the VOCASET width statistics. These four landmarks do not measure
anterior--posterior protrusion or constitute a dense mouth-surface evaluation.

\paragraph{CSV schema.}
The per-frame curve files should use the same schema for both original and retargeted
meshes:
\begin{center}
\small\texttt{dataset, subject\_id, sequence\_id, frame\_idx, opening, width, aperture}.
\end{center}
The preservation script merges original and retargeted curves using
\texttt{dataset}, \texttt{subject\_id}, \texttt{sequence\_id}, and \texttt{frame\_idx}.

\section{Additional Qualitative Retargeting Examples}
\label{sec:supp_qualitative}

\begin{figure}[H]
  \centering
  \includegraphics[width=0.72\linewidth]{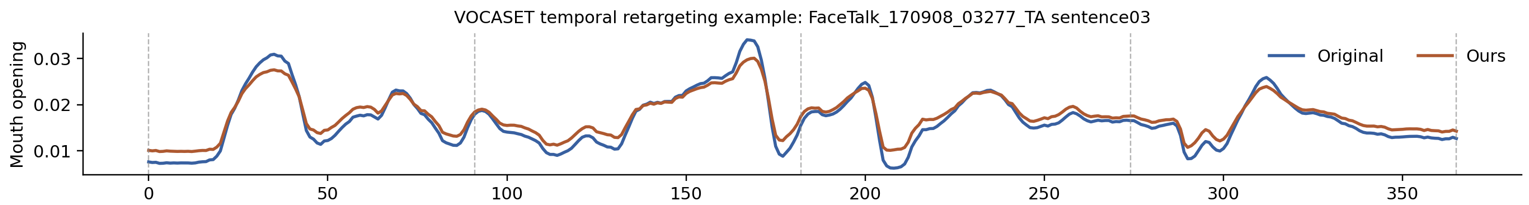}

  \vspace{2mm}
  \includegraphics[width=0.72\linewidth]{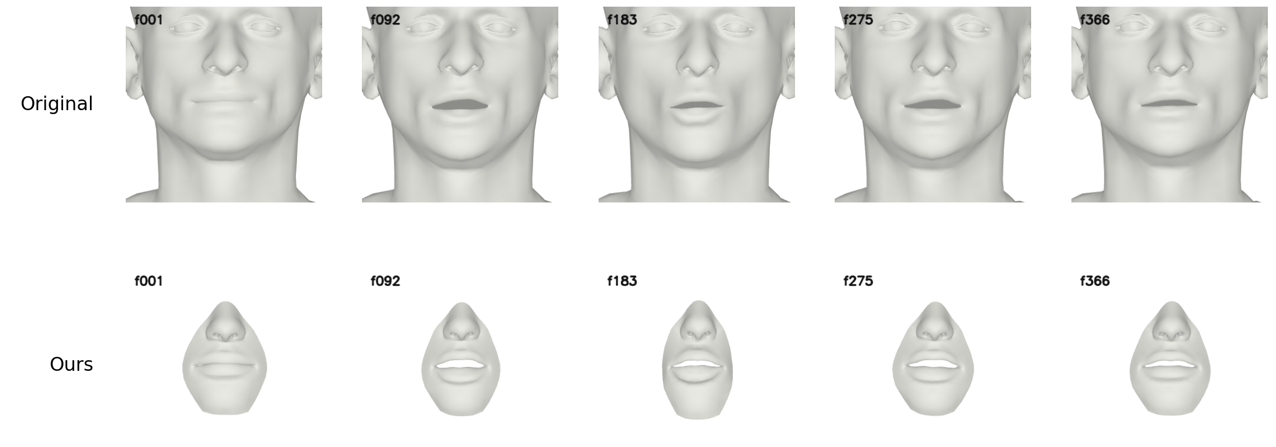}

  \caption{VOCASET temporal retargeting example. The top panel compares the
  mouth-opening trajectory of the original and retargeted sequence, and the lower
  panel shows representative frames at selected time steps.}
  \label{fig:supp_retargeting_vocaset_temporal}
\end{figure}

\begin{figure}[H]
  \centering
  \includegraphics[width=0.90\linewidth]{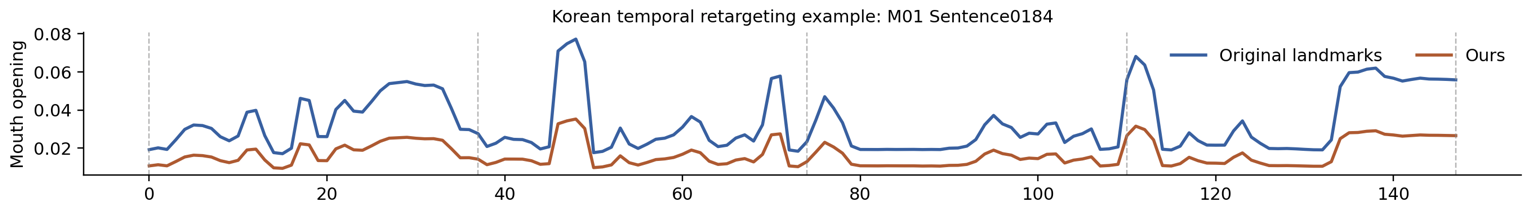}

  \vspace{2mm}
  \includegraphics[width=0.90\linewidth]{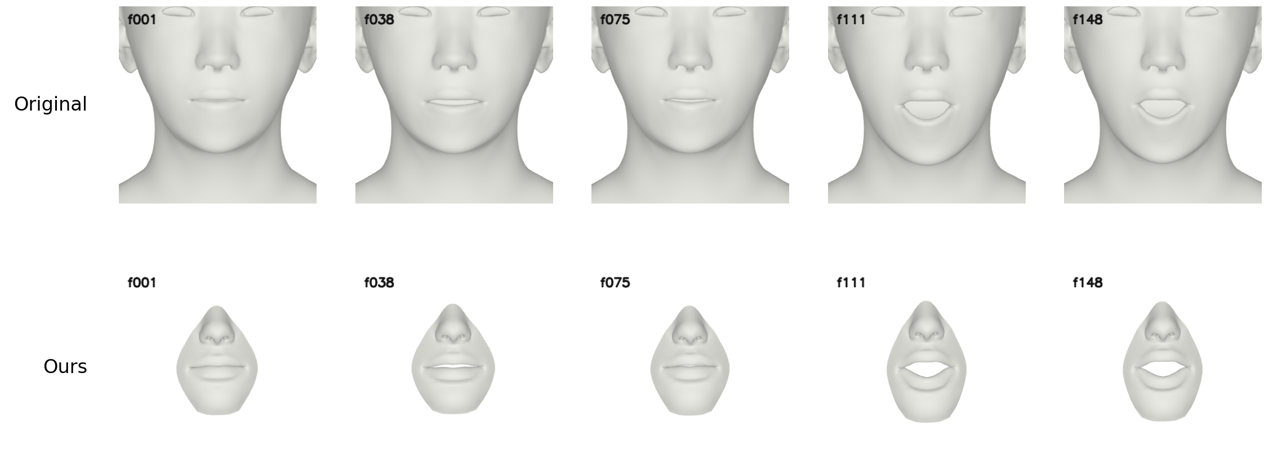}

  \caption{Korean temporal retargeting example. The top panel compares the
  mouth-opening trajectory of the original landmark sequence and the retargeted
  unified-template sequence, and the lower panel shows representative frames at
  selected time steps.}
  \label{fig:supp_retargeting_korean_temporal}
\end{figure}

In addition to the representative qualitative examples in the main paper, we provide
additional temporal retargeting examples from both VOCASET and Korean data.
Figures~\ref{fig:supp_retargeting_vocaset_temporal} and
\ref{fig:supp_retargeting_korean_temporal} complement the extended preservation
statistics in Table~\ref{tab:supp_mouth_preservation}. The supplementary PDF includes
four image panels in total: a VOCASET mouth-opening trajectory, VOCASET representative
frames, a Korean mouth-opening trajectory, and Korean representative frames. We group
each trajectory with its corresponding frame contact sheet so that the temporal curve
and selected frames can be read together while keeping labels and mesh details readable.

\end{document}